\documentclass{article}
\usepackage{ijcai22}

\usepackage{times}

\usepackage{soul}
\usepackage{url}
\usepackage[hidelinks]{hyperref}
\usepackage[utf8]{inputenc}
\usepackage[small]{caption}
\usepackage{graphicx}
\usepackage{amsmath}
\usepackage{booktabs}
\usepackage{amsmath}
\usepackage{amssymb}
\usepackage{amsthm}
\usepackage{arydshln}
\usepackage{listings}
\usepackage{subcaption}
\usepackage{multirow}
\usepackage{multicol}
\usepackage{tikz}
\usetikzlibrary{arrows,automata,shapes}
\newcommand{\tup}[1]{{\langle #1 \rangle}}
\usepackage{xcolor}

\newtheorem{definition}{Definition}

\title{Computing Programs for Generalized Planning as Heuristic Search \\(Extended Abstract)}

\author{
Javier Segovia-Aguas$^1$ \and
Sergio Jim\'enez$^2$\And
Anders Jonsson$^1$\\
\affiliations
$^1$Dept. Information and Communication Technologies, Universitat Pompeu Fabra, Spain\\
$^2$Valencian Research Institute for Artificial Intelligence, Universitat Polit\`ecnica de Val\`encia, Spain\emails
javier.segovia@upf.edu,
serjice@dsic.upv.es,
anders.jonsson@upf.edu
}

\begin{document}

\maketitle

\begin{abstract}
Although {\em heuristic search} is one of the most successful approaches to classical planning, this planning paradigm does not apply straightforwardly to {\em Generalized Planning} (GP).  This paper adapts the {\em planning as heuristic search} paradigm to the particularities of GP, and presents the first native heuristic search approach to GP. First, the paper defines a program-based solution space for GP that is independent of the number of planning instances in a GP problem, and the size of these instances. Second, the paper defines the {\sc BFGP} algorithm for GP, that implements a best-first search in our program-based solution space, and that is guided by different evaluation and heuristic functions.
\end{abstract}

\section{Introduction}
{\em Heuristic search} is one of the most successful approaches to classical planning~\cite{bonet2001planning,hoffmann2003metric,Helmert:FD:JAIR06,lipovetzky2017best}. Unfortunately, it is not straightforward to adopt state-of-the-art search algorithms and heuristics from classical planning to {\em Generalized Planning} (GP). The {\em planning as heuristic search} approach traditionally addresses the computation of sequential plans implementing a grounded state-space search. GP requires however reasoning about the synthesis of algorithm-like solutions that, in addition to action sequences, contain branching and looping constructs~\cite{Winner03distill:learning,Levesque:GPlanning:IJCAI11,Zilberstein:Gplanning:icaps11,Giacomo:FSM:ICAPS13,segovia2016generalized,illanes2019generalized,frances2021learning}. Furthermore, GP aims to synthesize solutions that generalize to a (possibly infinite) set of planning instances. The domain of the state variables may then be large, making  unfeasible the grounding traditionally implemented by of off-the-shelf classical planners. 

This paper adapts the {\em planning as heuristic search} paradigm to the particularities of GP, and presents the first native heuristic search approach to GP\footnote{This is an extended abstract of the ``Generalized Planning as Heuristic Search'' paper that appeared at ICAPS 2021 conference.}. Given a GP problem, that comprises an input set of classical planning instances from a given domain,  our {\em GP as heuristic search} approach computes an algorithm-like plan that solves the full set of input instances. The contribution of the paper is two-fold:
\begin{itemize}
	\item {\em A tractable solution space for GP}. We leverage the computational models of the {\em Random-Access Machine}~\cite{skiena1998algorithm} and the {\em Intel x86} FLAGS register~\cite{dandamudi2005installing} to define an innovative program-based solution space that is independent of the number of input planning instances in a GP problem, and the size of these instances (i.e. the number of state variables and their domain size).
	\item {\em A heuristic search algorithm for GP}. We present the {\sc BFGP} algorithm that implements a best-first search in our  solution space for GP.	We also define several evaluation and heuristic functions to guide {\sc BFGP}; evaluating these functions does not require to ground states/actions in advance, so they allow to addressing GP problems where state variables have large domains (e.g. integers). 
\end{itemize}

The paper is structured as follows. First we formalize the classical planning model. Then we show how to extend this model with a {\em Random-Access Machine} (RAM) and formalize GP with {\em planning programs}, our representation formalism for GP solutions. Last, we describe the implementation of our {\em GP as heuristic search} approach and report results on its empirical performance. More details on the {\em GP as heuristic search} approach can be found in~\citeauthor{segovia2021generalized}~[\citeyear{segovia2021generalized}].


\section{Classical Planning}
Let $X$ be a set of {\em state variables}, each $x\in X$ with domain $D_x$. A {\em state} is a total assignment of values to the set of state variables. For a variable subset $X'\subseteq X$, let $D[X']=\times_{x\in X'} D_x$ denote its joint domain. The state space is then $S=D[X]$. Given a state $s\in S$ and a subset of variables $X'\subseteq X$, let $s_{|X'}=\tup{x_i=v_i}_{x_i\in X'}$ be the {\em projection} of $s$ onto $X'$ i.e.~the partial state defined by the values that $s$ assigns to the variables in $X'$. The {\em projection} of $s$ onto $X'$ defines the subset of states $\{s \mid s \in S, s_{|X'}\subseteq s\}$ that are consistent with the corresponding partial state.

Let $A$ be a set of deterministic actions. An action $a\in A$ has an associated set of variables $par(a)\subseteq X$, called {\em parameters}, and is characterized by two functions: an {\em applicability function} $\rho_a: D[par(a)] \rightarrow \{0,1\}$, and a {\em successor function} $\theta_a: D[par(a)]\rightarrow D[par(a)]$. Action $a$ is applicable in a given state $s$ iff $\rho_a(s_{|par(a)})$ equals $1$, and results in a {\em successor} state $s'=s\oplus a$, that is built replacing the values that $s$ assigns to variables in $par(a)$ with the values specified by $\theta_a(s_{|par(a)})$.  

A {\em classical planning instance} is a tuple $P=\tup{X,A,I,G}$, where $X$ is a set of state variables, $A$ is a set of  actions, $I\in S$ is an initial state, and $G$ is a goal condition on the state variables that induces the subset of {\em goal states} 
$S_G = \{s \mid s \vDash G, s \in S\}$. Given $P$, a {\em plan} is an action sequence $\pi=\tup{a_1, \ldots, a_m}$ whose execution induces a {\em trajectory} $\tau=\tup{s_0, a_1, s_1, \ldots, a_m, s_m}$ such that, for each $1\leq i\leq m$, $a_i$ is applicable in $s_{i-1}$ and results in the successor state $s_i=s_{i-1}\oplus a_i$. A plan $\pi$ {\em solves} $P$ if the execution of $\pi$ in $s_0=I$ finishes in a goal state, i.e.~$s_m\in S_G$. 

\section{Generalized Planning as Heuristic Search}
This work builds on top of the inductive formalism for GP, where a GP problem is a set of classical planning instances with a common structure. Here we describe our heuristic search approach to GP.

\subsection{Classical Planning with a RAM}
To define a tractable solution  space  for  GP, that is  independent  of  the  number  (and  domain  size)  of  the  planning  state  variables, we  extend the classical planning model with: (i), a set of pointers over the state variables (ii), their primitive operations and (iii), two Boolean (the {\em zero} and {\em carry} FLAGS) to store the result of the primitive operations over pointers. 

Formally a {\em pointer} is a finite domain variable $z\in Z$ with domain $D_z=[0..|X|)$. To formalize the primitive operations over pointers we leverage the notion of the RAM; the RAM is an abstract computation machine, that is polynomially equivalent to a Turing machine, and that enhances a multiple-register {\em counter machine} with indirect memory addressing  ~\cite{boolos2002computability}. The indirect memory addressing of the RAM enables the definition of programs that access an unbounded number of state variables. Let $z\in Z$ be a pointer over the state variables, and $*z$ the content of that pointer, our {\em GP as heuristic search}
implements the following primitive operations over pointers: $\{{\tt\small inc}(z_1)$, ${\tt\small dec}(z_1)$, ${\tt\small cmp}(z_1,z_2)$, ${\tt\small cmp}(*z_1,*z_2)$, ${\tt\small set}(z_1,z_2)$ $| \; z_1,z_2 \in Z\}$. Respectively, these primitive operations increment/decrement a pointer, compare two pointers (or their content), and set the value of a pointer $z_1$ to another pointer $z_2$. Each primitive operation also updates two Boolean $Y=\{y_z,y_c\}$, the {\em zero} and {\em carry} FLAGS, according to the result (denoted here by {\tt\small res}) of that primitive operation:
\begin{footnotesize}
\begin{align*}
 inc(z_1) &\implies res := z_1 + 1,\\
 dec(z_1) &\implies res := z_1 - 1,\\
 cmp(z_1,z_2) &\implies res := z_1 - z_2,\\
 cmp(*z_1,*z_2) &\implies res := *z_1 - *z_2,\\ 
 set(z_1,z_2) &\implies res := z_2,\\
 y_z &:= ( res == 0 ),\\
 y_c &:= ( res > 0 ).
\end{align*}
\end{footnotesize}

Given a classical planning instance $P=\tup{X,A,I,G}$, its extension with a RAM of $|Z|$ pointers and two FLAGS is the classical planning instance $P_Z=\tup{X_Z,A_Z,I_Z,G}$, where the set of the state variables is  extended with the FLAGS and the pointers, $X_Z=X\cup Y\cup Z$. The set of actions $A_Z$ comprises the primitive pointer operations and the original actions $A$, but replacing their parameters by pointers in $Z$. The initial state $I_Z$ is extended to set the FLAGS to False, and the pointers to zero (by default). The goals of $P_Z$ are the same as those of the original instance. An extended instance $P_Z$ preserves the solution space of the original instance $P$~\cite{segovia2021generalized}.

\subsection{Generalized Planning with a RAM}
A GP problem is a set of classical planning instances with a common structure. In this work the common structure is given by the RAM extension; it provides a set of different classical planning instances with a common set of FLAGS, pointers, and actions defined over those pointers.

\begin{definition}[GP problem]
	\label{def:gp-problem}
A {\em GP problem} is a set of $T$ classical planning instances $\mathcal{P}=\{P_{Z}^1,\ldots,P_{Z}^{T}\}$ that share the same subset of state variables $\{Y\cup Z\}$, and actions $A_Z$, but may differ in their state variables, initial state, and goals. Formally, $P_{Z}^1=\tup{X_{Z}^{1},A_Z,I_{Z}^{1},G_1}, \ldots, P_{Z}^{T}=\tup{X_{Z}^{T},A_Z,I_{Z}^{T},G_T}$ where $\forall_{t} \{Y\cup Z\}\subset X_{Z}^{t}$, {\small $1\leq t\leq T$}.	
\end{definition}

Representations of GP solutions range from {\em programs}~\cite{Winner03distill:learning,celorrio2015computing,segovia2019computing} and {\it generalized policies}~\cite{Geffner:Gpolicies:AppliedI04}, to {\em finite state controllers}~\cite{Geffner:FSM:AAAI10,segovia2019computing} or {\em formal grammars} and {\em hierarchies} ~\cite{nau:shop2:JAIR03,segovia2017generating}. Each representation has its own expressiveness capacity, as well as its own computation complexity. We can however define a common condition under which a generalized plan is considered a solution to a GP problem~\cite{jimenez2019review}. First, let us define $exec(\Pi,P)=\tup{a_1, \ldots, a_m}$ as the sequential plan produced by the execution of a generalized plan $\Pi$, on a classical planning instance $P$.

\begin{definition}[GP solution]
	\label{def:gp-solution}
	A {\em generalized plan} $\Pi$ solves a GP problem $\mathcal{P}$ iff for every classical planning instance $P_t\in \mathcal{P}$, {\small $1\leq t\leq T$}, it holds that $exec(\Pi,P_t)$ solves $P_t$.
\end{definition}

In this work we represent GP solutions as {\em planning programs}~\cite{segovia2019computing}. A {\em planning program} is a sequence of $n$ instructions  $\Pi=\tup{w_0,\ldots,w_{n-1}}$, where each instruction $w_i\in \Pi$ is associated with a {\em program line} {\small $0\leq i< n$} and is either: 
\begin{itemize}
	\item A {\em planning action} $w_i\in A$.
	\item A {\em goto instruction} $w_i=\mathsf{go}(i',!y)$, where $i'$ is a program line $0\leq i'<i$ or $i+1<i'< n$, and $y$ is a proposition.
	\item A {\em termination instruction} $w_i=\mathsf{end}$. The last instruction of a program $\Pi$ is always  $w_{n-1}=\mathsf{end}$.
\end{itemize}

The execution model for a planning program  is a {\em program state} $(s,i)$, i.e.~a pair of a planning state $s\in S$ and program line $0\leq i< n$. Given a program state $(s,i)$, the execution of a programmed instruction $w_i$ is defined as:
\begin{itemize}
	\item If $w_i\in A$, the new program state is $(s',i+1)$, where $s'=s\oplus w_i$ is the {\em successor} when applying $w_i$ in $s$.
	\item If $w_i=\mathsf{go}(i',!y)$, the new program state is $(s,i+1)$ if $y$ holds in $s$, and $(s,i')$ otherwise\footnote{We adopt the convention of jumping to line $i'$ whenever $y$ is {\em false}, inspired by jump instructions in the {\em Random-Access Machine} that jump when a register equals zero.}. Proposition $y$ can be the result of an arbitrary expression on the state variables~\cite{lotinac2016automatic}. 
	\item If $w_i=\mathsf{end}$, program execution terminates. 
\end{itemize}

To execute a planning program $\Pi$ on a planning instance $P$, the initial program state is set to $(I,0)$, i.e.~the initial state of $P$ and the first program line of $\Pi$. A planning program $\Pi$ {\em solves} $P$ iff the execution terminates in a program state $(s,i)$ that satisfies the goal condition, i.e.~$w_i=\mathsf{end}$ and $s\in S_G$.

{\bf Example}. Figure~\ref{fig:cp-example} shows the initial state and goals of two classical planning instances, $P_1=\tup{X,A,I_1,G_1}$ and $P_2=\tup{X,A,I_2,G_2}$, for reversing two lists. Both instances can be defined with a set of state variables $X=\{x_i\}_1^l$, where an integer is assigned to every $x_i$ and $l$ is the list length, and a set of $swap(x_i,x_j)$ actions that swap the content of two state variables. An example solution plan for $P_1$ is  $\pi_1=\tup{swap(x_1,x_6), swap(x_2,x_5), swap(x_3,x_4)}$ while $\pi_2=\tup{swap(x_1,x_5), swap(x_2,x_4)}$ is a solution plan for $P_2$. This set of two classical planning instances can be extended with a RAM, and the resulting set $\mathcal{P}=\{P_{Z}^1,P_{Z}^2\}$ is an example GP problem. Figure~\ref{fig:lists} shows the generalized plan, represented as a planning program with six lines and two pointers $Z=\{z_1,z_2\}$. 
Initially, the first list element is pointed by $z_1$ (default) and the last element by $z_2$. The elements pointed by $z_1$ and $z_2$ are swapped in Line 0, and the pointers are moved forward and backward once respectively (Lines 1-2). Then, $z_2$ and $z_1$ are compared in Line 3, and the FLAGS are updated as follows: $y_z= ((z_2-z_1)==0)$ and $y_c = ((z_2-z_1)>0)$. This is repeated until $\neg(\neg y_z \wedge \neg y_c)$ is falsified, which only occurs when $z_2$ is smaller than $z_1$, thus reversing any list, no matter its length or content.

\begin{figure}[t]
	\centering
	\begin{tikzpicture}
	\draw[draw=black,step=0.5cm] (0.0,0.0) grid (2.5,0.5);
	\draw[draw=black,step=0.5cm] (4.0,0.0) grid (6.5,0.5);
	\draw[draw=black] (4.0,0.0) -- (4.0,0.5);
	\draw[draw=black,step=0.5cm] (0.0,1.0) grid (3.0,1.5);
	\draw[draw=black,step=0.5cm] (4.0,1.0) grid (7.0,1.5);
	\draw[draw=black] (4.0,1.0) -- (4.0,1.5);
	\draw[draw=black] (0.0,1.0) -- (3.0,1.0);
	\draw[draw=black] (4.0,1.0) -- (7.0,1.0);
	\draw[->,black] (2.6,0.25) -- (3.9,0.25);
	\draw[->,black] (3.1,1.25) -- (3.9,1.25);
	\node at (1.5,2.0) {\textbf{Initial State}};
	\node at (5.5,2.0) {\textbf{Goal State}};
	\node at (-0.5,0.25) {$P_2$};    
	\node at (0.25,0.25) {3};
	\node at (0.75,0.25) {2};
	\node at (1.25,0.25) {1};
	\node at (1.75,0.25) {5};
	\node at (2.25,0.25) {4};
	\node at (4.25,0.25) {4};
	\node at (4.75,0.25) {5};
	\node at (5.25,0.25) {1};
	\node at (5.75,0.25) {2};
	\node at (6.25,0.25) {3};
	\node at (-0.5,1.25) {$P_1$};        
	\node at (0.25,1.25) {6};
	\node at (0.75,1.25) {3};
	\node at (1.25,1.25) {4};
	\node at (1.75,1.25) {2};
	\node at (2.25,1.25) {5};
	\node at (2.75,1.25) {1};
	\node at (4.25,1.25) {1};
	\node at (4.75,1.25) {5};
	\node at (5.25,1.25) {2};
	\node at (5.75,1.25) {4};
	\node at (6.25,1.25) {3};
	\node at (6.75,1.25) {6};
	\end{tikzpicture}
	\caption{Classical planning instances for reversing the content of two lists by swapping their elements.}
	\label{fig:cp-example}
\end{figure}
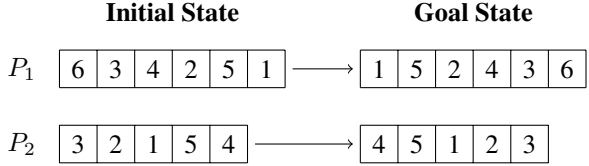

\begin{figure}[t]
\footnotesize
\begin{lstlisting}[mathescape]
            REVERSE
            0. swap($z_1$,$z_2$)
            1. inc($z_1$)
            2. dec($z_2$)
            3. cmp($z_2$,$z_1$)
            4. goto(1,$\neg$($\neg y_z\wedge\neg y_c$))
            5. end
\end{lstlisting}
	\caption{{\em Generalized plan} for reversing a list, no matter its length.}
	\label{fig:lists}
\end{figure}

\subsection{The {\sc BFGP} algorithm for Generalized Planning}
Given a GP problem $\mathcal{P}=\{P_1,\ldots,P_T\}$, a number of program lines $n$, and a number of pointers $|Z|$, the {\sc BFGP} algorithm outputs a planning program $\Pi$ that solves every classical planning instance $P_t\in \mathcal{P}$, {\small $1\leq t\leq T$}. Otherwise {\sc BFGP} reports that there is no solution within the given number of program lines and pointers. 

{\bf Search space.}  {\sc BFGP} searches in the space of planning programs with $n$ program lines, and $|Z|$ pointers, that can be built with the shared set of actions $A_Z$, and goto instructions that are exclusively conditioned on the value of FLAGS $Y=\{y_z,y_c\}$. Since only the primitive operations over pointers update FLAGS $Y$, we have an observation space of $2^{|Y|}\times 2|Z|^2$ state observations implemented with only $|Y|$ Boolean variables. The four joint values of $\{y_z,y_c\}$ model then a large space of observations, including $=\,$0, $\neq\,$0, $<0, >0, \leq 0, \geq 0$ as well as relations $=, \neq, <, >, \leq, \geq$ on variable pairs.

{\bf Search algorithm.} {\sc BFGP} implements a Best First Search (BFS) that starts with an empty planning program. To generate a tractable set of successor nodes, child nodes in the search tree are restricted to planning programs that result from programming the $PC^{MAX}$ line (i.e. the maximum line reached after executing the current program on the classical planning instances in $\mathcal{P}$). This procedure for successor generation guarantees that duplicate successors are not generated. {\sc BFGP} is a {\em frontier search} algorithm, meaning that, to reduce memory requirements, {\sc BFGP} stores only the open list of generated nodes, but not the closed list of expanded nodes~\cite{korf2005frontier}. BFS sequentially expands the best node in a priority queue (aka {\em open list}) sorted by an evaluation/heuristic function. If the planning program $\Pi$ solves all the instances $P_t\in \mathcal{P}$, then search ends, and $\Pi$ is a valid solution for the GP problem $\mathcal{P}$. 

\begin{table*}[ht!]
	\centering
	\begin{tiny}    
		\begin{tabular}{|l|c||c|c|c|c||c|c|c|c||c|c|c|c|} \hline
			\multirow{2}{*}{\textbf{Domain}} & \multirow{2}{*}{$n,|Z|$} & \multicolumn{4}{|c||}{$f_1$} & \multicolumn{4}{|c||}{$f_2$} & \multicolumn{4}{|c|}{$f_3$}  \\\cline{3-14}
			& & Time & Mem. & Exp. & Eval.  & Time & Mem. & Exp. & Eval.  & Time & Mem. & Exp. & Eval.  \\\hline
			T. Sum & 5, 2 & 0.24 & 4.2 & 4.8K & 5.8K & 0.30 & {\bf 3.8} & 6.4K & 6.4K & 0.13 & 3.9 & 2.2K & 2.8K \\
			Corridor & 7, 2 & 3.04 & 6.6 & 12.4K & 26.7K & {\bf 0.41} & {\bf 3.8} & {\bf 2.2K} & {\bf 2.2K} & 3.66 & 4.2 & 24.2K & 25.9K \\
			Reverse & 7, 3 & 82 & 61 & 0.28M & 0.57M & 181 & {\bf 4.0} & 0.95M & 0.95M & 170 & 23 & 0.75M & 0.84M \\
			Select & 7, 3 & 198 & 110 & 0.83M & 1.10M & 27 & {\bf 3.9} & 0.12M & 0.12M & 76.49 & 11 & 0.34M & 0.38M \\
			Find & 7, 3 & 195 & 175 & 0.50M & 1.36M & 271 & {\bf 4.0} & 1.46M & 1.46M & {\bf 86} & 12 & {\bf 0.41M} & {\bf 0.45M} \\
			Fibonacci & 8, 3 & {\bf 496} & 922 & {\bf 2.48M} & {\bf 6.79M} & 1,082 & {\bf 3.9} & 11.8M & 11.8M & TO & - & - & - \\
			Gripper & 8, 4 & TO & - & - & - & 3,439 & {\bf 4.1} & 19.9M & 19.9M  & TO & - & - & - \\
			Sorting & 9, 3 & TO & - & - & - & TO & - & - & -  & {\bf 3,143} & {\bf 711} & {\bf 19.5M} & {\bf 22.9M} \\\hline
		    \multicolumn{2}{|c||}{Average} & 162 & 213 & 0.68M & 1.64M & 714 & 3.9 & 4.89M & 4.89M & 580 & 128 & 3.51M & 4.09M \\\hline
		\end{tabular}
		\begin{tabular}{|l|c||c|c|c|c||c|c|c|c||c|c|c|c|} \hline
			\multirow{2}{*}{\textbf{Domain}} & \multirow{2}{*}{$n,|Z|$} &  \multicolumn{4}{|c||}{$h_4$} & \multicolumn{4}{|c||}{$h_5$} & \multicolumn{4}{|c|}{$f_6$} \\\cline{3-14}
			& & Time & Mem. & Exp. & Eval.  & Time & Mem. & Exp. & Eval.  & Time & Mem. & Exp. & Eval.  \\\hline
			T. Sum & 5, 2 & 0.10 & {\bf 3.8} & 1.6K & 1.6K & {\bf 0.09} & 3.9 & {\bf 1.2K} & {\bf 1.4K} & 0.45 & 4.8 & 7.4K & 7.4K \\
			Corridor & 7, 2 & 1.09 & 3.9 & 5.3K & 5.4K & 5.29 & 4.1 & 30.3K  & 31.3K & 6.16 & 7.6 & 35.3K & 35.3K \\
			Reverse & 7, 3 & 205 & 4.2 & 0.88M & 0.88M & {\bf 1.46} & 4.2 & {\bf 4.9K} & {\bf 6.3K} & 369 & 230 & 1.57M & 1.65M \\
			Select & 7, 3 & {\bf 0.80} & {\bf 3.9} & {\bf 3.0K} & {\bf 4.2K} & 94 & 5.7 & 0.34M & 0.35M & 255 & 155 & 1.06M & 1.14M \\
			Find & 7, 3 & 415 & 4.4 & 1.76M & 1.76M & 140 & 7.0 & 0.58M & 0.59M & 423 & 244 & 1.76M & 1.77M \\
			Fibonacci & 8, 3 & TO & - & - & - & 1,500 & 120 & 11.3M & 11.8M & TO & - & - & - \\
			Gripper & 8, 4  & TO & - & - & - & {\bf 83} & 5.5 & {\bf 0.34M} & {\bf 0.35M} & TO & - & - & - \\
			Sorting & 9, 3 & TO & - & - & - & TO & - & - & -  & TO & - & - & - \\\hline
		    \multicolumn{2}{|c||}{Average} & 125 & 4.0 & 0.53M & 0.53M & 260 & 21 & 1.80M & 1.88M & 211 & 128 & 0.89M & 0.92M \\\hline
		\end{tabular}
	\end{tiny}
	\caption{ We report the number of program lines $n$, and pointers $|Z|$ per domain, and for each evaluation/heuristic function, CPU (secs), memory peak (MBs), and the numbers of expanded and evaluated nodes. TO stands for Time-Out ($>$1h of CPU). Best results in bold.}
	\label{tab:heuristics}
\end{table*}

\begin{table}[t]
\begin{tiny} 
    \centering
    \begin{tabular}{|l||c|c|c|c|} \hline
        \multirow{2}{*}{\bf Dom.} & \multicolumn{4}{|c|}{BFGP$(f_1,h_5)$ / BFGP$(h_5,f_1)$}\\\cline{2-5} 
          & T.  & M. & Exp. & Eval.  \\\hline
            T. Sum & 0.2/{\bf 0.1} & 4.3/{\bf 3.8} & 2.8K/{\bf 1.1K} & 4.8K/{\bf 1.4K}\\
			Corr. & {\bf 3.2}/4.5 & 6.6/{\bf 5.9} & {\bf 6.5K}/26.0K & {\bf 21.6K}/27.5K \\
			Rev.  & 63/{\bf 1.4} & 52/{\bf 4.7} & 81.9K/{\bf 3.7K} & 0.3M/{\bf 7.7K}\\
			Sel.  & 203/{\bf 80}  & 110/{\bf 7.0} & 0.6M/{\bf 0.3M} & 0.9M/{\bf 0.3M} \\
			Find  & 313/{\bf 162} & 176/{\bf 14} & 0.9M/{\bf 0.7M} & 1.5M/{\bf 0.7M} \\
			Fibo.  & 528/{\bf 22} & 828/{\bf 32} & 1.4M/{\bf 75K} & 5.3M/{\bf 0.2M} \\
			Grip.  & TO/{\bf 6.9} & -/{\bf 10} & -/{\bf 5.8K} & -/{\bf 37.3K}   \\
			Sort.  & TO/{\bf 713} & -/{\bf 730} & -/{\bf 4.4M} & -/{\bf 4.5M}  \\\hline
    \end{tabular} 
    \caption{CPU time (secs), memory peak (MBs), num. of expanded and evaluated nodes. Best results in bold.}
    \label{tab:h-combined}
    \end{tiny}
\end{table}

\begin{table}[t]
	\centering
	\begin{tiny}
		\begin{tabular}{|l|r||r|r||r|r|}\hline
			\textbf{Dom.}  & Inst. & Time$_\infty$ & Mem$_\infty$ & Time & Mem \\\hline
			T. Sum & 44,709 & 1,066.74 & 53MB & \textbf{574.08} & \textbf{47MB} \\  
			Corr. & 1,000 & 0.23 & 5.0MB & \textbf{0.15} & {\bf 4.7MB} \\
			Rev. & 50 & 37.96 & 5.2GB & \textbf{2.70} & \textbf{0.3GB} \\
			Sel. & 50 & 144.75 & 19.6GB & \textbf{2.29} & \textbf{33MB} \\
			Find & 50 & 114.55 & 19.6GB & \textbf{2.12} & \textbf{33MB}\\
			Fibo. & 33 & \textbf{0.00} & 4.2MB & \textbf{0.00} & \textbf{3.9MB} \\
			Grip. & 1,000 & 2.71 & 0.1GB & \textbf{1.65} & \textbf{0.1GB} \\
			Sort. & 20 & 272.06 & 15.2GB & \textbf{52.04} & \textbf{3.8MB} \\\hline
		\end{tabular}
	\end{tiny}
	\caption{Validation set, CPU-time (secs) and memory peak for program validation, with/out {\em infinite program} detection.}
	\label{tab:validation}
\end{table}

{\bf Evaluation functions.} {\sc BFGP} exploits two different sources of information to guide the search in the space of candidate planning programs:
\begin{itemize}
	\item {\em The program structure}. These are evaluation functions computed in linear time in the size of program $\Pi$. 
		\begin{itemize}
			\item$f_1(\Pi)$, the number of {\em goto} instructions in $\Pi$.
			\item$f_2(\Pi)$, number of {\em undefined} program lines in $\Pi$.
			\item$f_3(\Pi)$, the number of repeated actions in $\Pi$.
		\end{itemize}
	
	\item {\em The program performance}. These functions assess the performance of $\Pi$ executing it on each of the classical planning instances $P_t\in\mathcal{P}$, {\small $1\leq t \leq T$}; the execution of a planning program on a classical planning instance is a deterministic procedure that requires no variable instantiation. If the execution of $\Pi$ on an instance $P_t\in \mathcal{P}$ fails, this means that the search node corresponding to the planning program $\Pi$ is a dead-end, and hence it is not added to the open list:
		\begin{itemize}
			\item$h_4(\Pi,\mathcal{P})=n-PC^{MAX}$, where $PC^{MAX}$ is the maximum program line that is eventually reached after executing $\Pi$ on all the  instances in $\mathcal{P}$. 
			\item $h_5(\Pi,\mathcal{P})=\sum_{P_t\in\mathcal{P}}\sum_{v\in G_t} (s_t[v]-G_t[v])^2$. This function accumulates, for each instance $P_t\in\mathcal{P}$, the {\em euclidean distance} of state $s_t$ to the goal state variables $G_t$. The state $s_t$ is obtained applying the sequence of actions  $exec(\Pi,P_t)$ to the initial state $I_t$ of that problem $P_t\in\mathcal{P}$. Computing $h_5(\Pi,P_t)$ requires that goals are specified as a partial state. Note that for Boolean variables the squared difference becomes a simple goal counter.
			\item $f_6(\Pi,\mathcal{P}) = \sum_{P_t\in{\cal P}} |exec(\Pi,P_t)|$, where $exec(\Pi,P_t)$ is the sequence of actions induced from executing the planning program $\Pi$ on the planning instance $P_t$. 
		\end{itemize}
\end{itemize}
All these functions are {\em cost functions} (i.e.~smaller values are preferred). Functions $h_4(\Pi,\mathcal{P})$ and $h_5(\Pi,\mathcal{P})$ are cost-to-go {\em heuristics}; they provide an estimate on how far a program is from solving the given GP problem.  Functions $h_4(\Pi,\mathcal{P})$, $h_5(\Pi,\mathcal{P})$, and $f_6(\Pi,\mathcal{P})$ aggregate several costs that could be expressed as a combination of different functions, e.g.~{\em sum}, {\em max}, average, weighted average, etc.

\section{Evaluation}
We evaluated the performance of our {\em GP as heuristic search} approach in eight domains~\cite{segovia2021generalized}. Experiments performed in an Ubuntu 20.04 LTS, with AMD® Ryzen 7 3700x 8-core processor $\times$ 16 and 32GB of RAM. 

Table~\ref{tab:heuristics} summarizes the results of {\sc BFGP} with our six different evaluation/heuristic functions (best results in bold): i) $f_2$ and $h_5$ exhibited the best coverage; ii) there is no clear dominance of a {\em structure} evaluation function, $f_2$ has the best memory consumption while $f_3$ is the only structural function that solves {\em Sorting}; iii) the {\em performance}-based function $h_5$ dominates $h_4$ and $f_6$. Interestingly, the base performance of {\sc BFGP} with a single evaluation/heuristic function is improved guiding {\sc BFGP} with a cost-to-go heuristic function and breaking ties with a structural evaluation function (or vice versa). Table~\ref{tab:h-combined} shows the performance of $BFGP(f_1,h_5)$ and its reversed configuration $BFGP(h_5,f_1)$ which actually resulted in the overall best configuration solving all domains. 

The solutions synthesized by {\sc BFGP} were successfully validated. Table~\ref{tab:validation} reports the CPU time, and peak memory, yield when running the solutions synthesized by $BFGP(h_5,f_1)$ on a validation set. The largest CPU-times and memory peaks correspond to the configuration that implements the detection of {\em infinite programs}, which requires saving states to detect whether they are revisited during execution. Skipping this mechanism allows to validate non-infinite programs faster~\cite{aguas2020generalized}.

\section{Conclusion}
We presented the first native heuristic search approach for GP that leverages solutions represented as automated planning programs. We believe this work builds a stronger connection between the two closely related areas of planning and {\em program synthesis}~\cite{gulwani2017program,alur2018search}. A wide landscape of effective techniques, coming from heuristic search and classical planning, promise to improve the base performance of our approach~\cite{segovia2022landmarks}. For instance, better estimates may be obtained by building on top of better informed planning heuristics~\cite{frances2017effective}. 

\section*{Acknowledgements}
\begin{footnotesize}
Javier Segovia-Aguas is supported by TAILOR, a project funded by EU H2020 research and innovation programme no. 952215. Sergio Jim\'enez is supported by Spanish grants RYC-2015-18009 and  TIN2017-88476-C2-1-R. Anders Jonsson is partially supported by Spanish grant PID2019-108141GB-I00.
\end{footnotesize}

\bibliographystyle{named}
\bibliography{ijcai22}

\end{document}